\documentclass[11pt]{article}
\usepackage{amssymb}
\usepackage{lipsum}
\usepackage[final]{acl}

\usepackage{times}
\usepackage{latexsym}
\usepackage{amsmath}
\usepackage{booktabs}
\usepackage[T1]{fontenc}
\usepackage[utf8]{inputenc}
\usepackage{microtype}
\usepackage{inconsolata}
\usepackage{graphicx}
\usepackage{url}

\title{Canonical Color as a Lens into Concept Decodability in Vision Encoders and VLMs}

\author{
  Xiaofu Chen$^{1}$ \quad
  Stella Frank$^{2}$ \quad
  Yova Kementchedjhieva$^{1}$ \\
  $^{1}$MBZUAI \quad
  $^{2}$Technical University of Denmark \\
  \texttt{\{xiaofu.chen, yova.kementchedjhieva\}@mbzuai.ac.ae} \\
  \texttt{stefra@dtu.dk}
}

\begin{document}
\maketitle

\begin{abstract}
Visual encoders construct a representation of the image input for Vision-Language models.
How much conceptual, as opposed to immediately visible, information does this representation contain?
We use canonical color as a controlled test case to ask whether vision encoders make canonical-color information linearly accessible, even when color is removed from the input image.
We construct a dataset of objects with canonical colors, and probe vision encoders for both color and object identity using color and grayscale images.
We find that canonical color remains decodable from grayscale images, and is tied to predicted object identity, indicating a conceptual link.
Extending this analysis to full VLMs, we find that VLM post-training can have a surprisingly large effect on color decodability in the vision encoder.
Overall, canonical color provides a usefully controllable lens for tracing object-level conceptual semantic information in vision encoders and VLMs.

\end{abstract}

\section{Introduction}

Humans associate object concepts with typical attributes, including canonical color. Such attributes are part of conceptual representations, and color can be especially diagnostic for some object categories \citep{mcrae2005semantic,tanaka1999color}. Human studies also show that object knowledge can affect perceived color \citep{hansen2006memory,witzel2011object}. Thus, knowing that bananas are yellow is different from seeing yellow pixels in a particular image. For text-only models, acquiring such typical knowledge can be difficult because text often states unusual rather than obvious properties, a problem known as reporting bias \citep{gordon2013reporting}. Multimodal semantic work has therefore used visual grounding to complement linguistic evidence \citep{bruni2014multimodal,kiela2014learning,silberer2014learning}. Vision encoders have direct access to object colors during training, but it remains unclear whether they encode canonical color as object-level conceptual knowledge or simply expose chromatic cues from the input during inference.

Recent work has begun to address this question with representation-level probing. Linear probes are widely used to study what information is accessible in intermediate representations, although their results depend on the probe and control setting \citep{alain2017understanding,hewitt2019designing,belinkov2022probing}. \citet{oneata2025seeing} show that frozen vision encoders encode many semantic attributes, including color. However, their stimuli are full-color RGB images, so color prediction may reflect visible pixels rather than canonical color knowledge. In VLMs, final answers can also be shaped by language priors, dataset bias, hallucination, or weak object--attribute binding \citep{goyal2017making,rohrbach2018object,li2023evaluating,yuksekgonul2023when}. \citet{golovanevsky2025pixels} study such conflicts with counterfactual images, e.g., a blue strawberry, but full-model behavior does not reveal where canonical color information is represented within the vision--language pipeline.

In this work, we use canonical color as a controlled probe of object-level semantic information in vision encoders and VLMs. 
Color is a useful case because, unlike many other visual attributes, it can be directly removed from the image input, by gray-scaling the image, while leaving the object structure intact.
Because canonical color is defined at the object-category level, its relation to object identity is part of the question we study, rather than a confound. We ask whether canonical color remains linearly accessible through object-level representations after visible color is removed, and whether object recognition alone accounts for its decodability.
In other words, we probe the encoder: How is a black-and-white banana like black-and-white daffodils and lemons?

Furthermore, if visual encoders represent canonical colors as part of their object concept representations, how is this information passed on to the language backbone in VLM architectures?
Does VLM training preserve the conceptual organization in the visual encoder, or is there a conceptual reorganization towards the language model?

To answer these questions, we construct a dataset of object classes annotated with basic canonical color labels and train linear probes on frozen representations. 
RGB inputs provide an upper-bound setting with visible color cues, while our \textbf{Grayscale} setting removes chromatic information and reduces simple luminance cues through grayscale conversion and histogram equalization. 
We validate this control with Visual-CounterFact~\citep{golovanevsky2025pixels} by comparing actual-color prediction from recolored images with canonical-color prediction from both original and counterfactual images. 
We then extend the probing logic to VLMs by comparing matched pre- and post-VLM vision towers and decoder-side visual-token representations.

Our results show that canonical color remains linearly decodable even after direct color cues are removed. Counterfactual validation indicates that this residual signal is better aligned with canonical object color than with actual surface color. We also find that canonical color is related to object-class information, but the two signals are only partially aligned. Finally, VLM post-training changes where object and color information are most linearly accessible: in some models, this information becomes less accessible in the standalone vision tower but reappears after the visual interface or inside decoder-side visual-token states.

\section{Related Work}
\label{sec:related_work}

Semantic feature norms describe object concepts through typical properties, including perceptual attributes such as color \citep{mcrae2005semantic,devereux2014centre, bannert2013decoding}. Recent work has used probing to ask whether pretrained representations encode such attributes. \citet{oneata2025seeing} probe image encoders, multimodally trained image encoders, and language-only models for perceptual, functional, and encyclopedic attributes. Our work follows this representation-level view, but focuses on canonical color as a controlled case study. Color is a special attribute because it is directly visible in RGB images: a successful color probe may rely on pixel-level chromatic cues rather than object-level canonical color.We therefore introduce input-side controls to remove chromatic information and further reduce the influence of luminance cues.

Linear probes are useful for tracing what information is accessible in frozen representations \citep{alain2017understanding}, but probe accuracy alone does not show that a model uses that information in its final predictions. Prior work therefore emphasizes careful probe design, control tasks, and cautious interpretation \citep{hewitt2019designing,belinkov2022probing}. We follow this view by treating probing as a measure of linear accessibility, and by validating our color-controlled setting with counterfactual images.

Work on VLMs further shows that model predictions can be shaped by language priors, dataset bias, hallucination, and weak object--attribute binding \citep{goyal2017making,rohrbach2018object,li2023evaluating,zhao2022vlchecklist,yuksekgonul2023when}. More directly related to color, \citet{tang-etal-2023-lemons} identify Concept Association Bias, where VLMs fill in strongly associated missing concepts across modalities. \citet{liang2025colorbench} show that color perception, reasoning, and robustness remain challenging for modern VLMs. \citet{golovanevsky2025pixels} further study conflicts between visual evidence and memorized priors using Visual CounterFact, where objects are recolored to non-canonical colors. Unlike these behavioral studies, we localize where object and canonical color information are linearly decodable across vision encoders, VLM-trained vision towers, visual interfaces, and decoder-side visual-token representations.

\section{Data and Probing Setup}
\label{sec:dataandprobing}
This section describes our controlled probing setup. We first construct a dataset of object--color pairs, where each object class is associated with a canonical color label. We then apply image-side controls to separate visible pixel color from canonical object color. Finally, we pass the images through frozen vision encoders, extract layer-wise representations, and train linear probes to predict canonical color and object identity from these representations.

\subsection{Canonical Color Dataset}

We construct a dataset of 708 object classes, each associated with one of 10 canonical-color labels, with a target of five images per class.

Our dataset extends the object--color pairs compiled by \citet{golovanevsky2025pixels} with additional pairs sourced from Wikidata. We retain entries whose structured \textit{has-color} property contains a single value, which provides the canonical-color label. To reduce the label-space complexity, we map these values to ten basic colors: black, blue, brown, gray, green, pink, purple, red, white, and yellow. The resulting object--color pairs are manually screened for validity. This step revealed that the data sourced from Wikidata was dominated by Pokémon characters. We downsample those to 50 classes, prioritizing colors of low frequency among the remaining classes (gray, brown, and red) to achieve better balance across colors. The resulting color distribution is shown in Figure~\ref{fig:color_distribution}---white is the most frequent label with 99 classes and gray the least frequent with 48.

For each class, we obtain five images by combining those linked from Wikidata, where available, with additional Google Search results, retaining up to five images that clearly depict the object and are consistent with its canonical color.

\subsection{Input-side Controls for Color Probing}
\label{sec:visual_leakage}

A probe that predicts the canonical color of an object from a vision encoder's representation may rely on color information that is directly present in the input image. RGB images contain real pixel-level color, so high probing accuracy in this setting does not by itself show that the encoder stores color as concept-level knowledge. %
We therefore treat RGB as an upper-bound setting and remove this direct color channel in the controlled settings.

To remove pixel color, we first convert each image to grayscale by using a single luminance channel and replicating it across three channels. We then apply per-image histogram equalization to reduce brightness-based cues that may still correlate with color. We refer to this processed input as \textbf{grayscaling} throughout the paper. %

\subsection{Probing Tasks and Evaluation}
\label{sec:probing_setup}

We probe the internal representations of frozen vision encoders to ask two questions. 
First, do these representations contain information that can predict an object's canonical color? 
Second, how does this color information relate to object-class information? 
To answer these questions, we use two linear probing tasks. 
The color probe predicts one of the 10 color classes in our dataset, while the object-class probe predicts the object identity among 708 object classes.

We compare the layer-wise trajectories of the two probes to test whether canonical color decodability is associated with category-level object semantics. 
Because the two tasks have different label spaces, 10 color classes versus 708 object classes, their absolute accuracies are not directly comparable. 
We therefore focus on how their accuracies change across layers and whether the two trends become aligned.

Canonical color labels are defined over object categories rather than over individual image instances. 
For example, the label ``yellow'' for an image containing a banana reflects the canonical color associated with the object category \textit{banana}. 
Thus, a color probe that predicts ``yellow'' for such an image may rely on representations that support category-level identification of the object as a banana, or on other image-level cues that correlate with canonical color. 
A color signal that appears when object-class recognition is still weak may reflect lower-level visual regularities or coarse object information. 
A color signal that strengthens together with object-class recognition is more likely to be linked to category-level object semantics.

\section{Controlled Probing of Canonical Color in Vision Encoders}
\label{result}

\subsection{Experimental Setup}
\label{sec:experimental_setup}

\paragraph{Vision Encoders.}
We evaluate five vision encoders that cover different supervision signals and architectural designs. CLIP~\citep{radford2021learning} and SigLIP~\citep{zhai2023sigmoid} are language-supervised models trained with image--text alignment objectives, so their visual representations are shaped by natural language supervision. DINOv2~\citep{oquab2023dinov2} and ViT-MAE~\citep{he2022masked} are self-supervised vision models, but they use different objectives: DINOv2 learns through self-distillation, whereas ViT-MAE learns by reconstructing masked image patches. We also include Swin-V2~\citep{liu2022swinv2}, a hierarchical vision transformer, to test whether the observed patterns hold beyond standard ViT-style backbones.
All models are used at the ViT-Base scale, or at the corresponding hierarchical scale for Swin-V2. Model repositories are listed in Appendix~\ref{sec:appendix}.

\paragraph{Features.}
For each input image, we cache hidden states from all transformer blocks and use mean-pooled patch features for all main probing results.
For Swin-V2, adjacent blocks within each stage are averaged to form 12 layer-indexed features, matching the ViT-Base models.

\paragraph{Probing.}
For each vision encoder, input setting, layer, and representation type, we train an independent linear probe on the cached image-level features. Each probe is an L2-regularized multinomial logistic regression classifier trained for up to 1000 optimization iterations. The vision encoders remain frozen throughout probing.

For color probing, we use 5-fold object-class-level cross-validation to avoid leakage across images of the same object class. We split the 708 object classes into five disjoint folds. In each fold, the probe is trained on all images from the training object classes and evaluated on the held-out object classes. At test time, we average the predicted probability vectors across all images of each test object class and take the argmax, producing one color prediction per object class. We report accuracy across the five folds.

For object-class recognition, we use a per-class image hold-out setting, since holding out entire object classes would remove the target labels from training. For each object class with at least two images, we hold out one image for testing and use the remaining images for training. Object classes with only one image are used for training only. This task tests whether different images of the same object class are grouped in the representation space.

\subsection{Validating the Color-controlled Setting}
\label{sec:input_control_validation}

\begin{figure}[t]
    \centering
    \includegraphics[width=\linewidth]{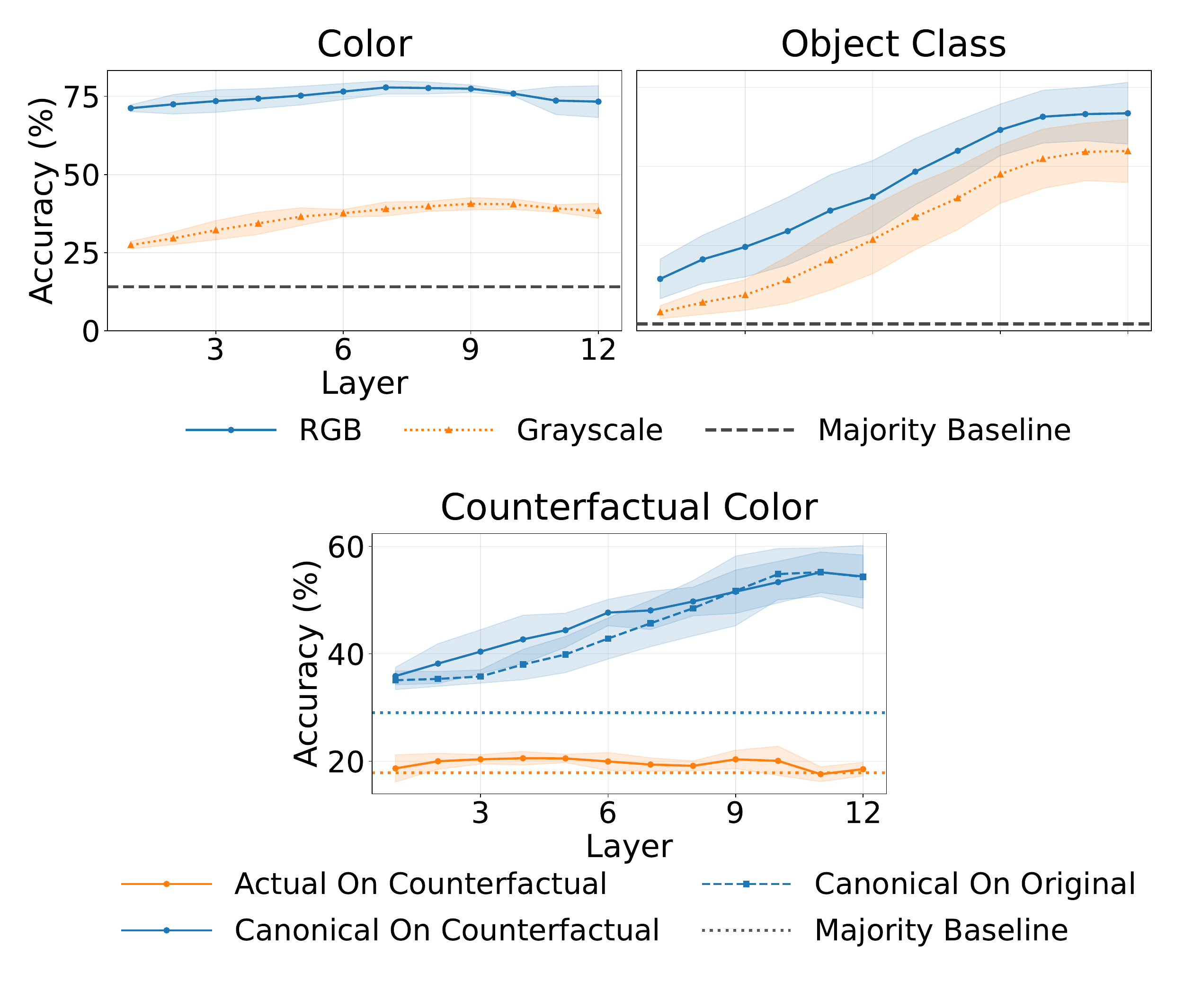}
    \caption{
    Validation of the grayscale control.
    Top left: color probing drops after visible color cues are removed.
    Top right: object-class recognition remains decodable.
    Bottom: No trace of the counterfactual color can be decoded from grayscale images, while canonical color can be decoded from both grayscale and counterfactual images.
    Dashed lines show majority baselines; shaded bands show $\pm 1$ standard deviation across encoders.
    }
    \label{fig:input_counterfactual_validation}
\end{figure}

Before using grayscale as our main controlled setting, we first validate what this preprocessing removes and what it preserves (Figure~\ref{fig:input_counterfactual_validation}). 
All experiments use patch-averaged representations from the same five frozen encoders used in the main experiments, with results averaged across encoders.

First, we observe that color-probing accuracy drops sharply from RGB to grayscale, confirming that RGB probing is strongly driven by visible pixel color. 
Second, object-class recognition also drops under grayscale, but remains clearly decodable and continues to improve with depth. 
Thus, grayscaling removes direct color cues while still preserving enough object-level structure for semantic probing.

The top two panels do not tell us whether the remaining color signal reflects the actual surface color in the image or the canonical color of the object. 
To separate these possibilities, we repeat the probing analysis on Visual-CounterFact, which contains paired original and recolored images of the same object~\citep{golovanevsky2025pixels}. 
We apply the same grayscale preprocessing to all images. 
The actual-color probe on counterfactual images stays close to its majority baseline, indicating that the counterfactual surface color is mostly removed. 
In contrast, the two canonical-color probes, one on counterfactual images and one on original images, are both well above their majority baseline and show similar layer-wise trends. 
The actual-color probe and the canonical-color probes use different label spaces. 
We therefore use the actual-color probe only as a check that the counterfactual surface color is not recoverable after grayscale preprocessing. 
The main validation comes from comparing the canonical-color probes on the original and the counterfactual images, which use the same canonical-color label space. 
Their similar layer-wise behavior shows that canonical color remains decodable from both counterfactual and original images after visible color cues are removed. 

Overall, these controls address two types of residual cues. First, grayscale conversion removes chromatic cues, histogram equalization reduces simple brightness cues, and Visual-CounterFact tests whether the counterfactual surface color can still be decoded after preprocessing. The actual-color probe remains near the majority baseline, while the canonical-color probes on the original and counterfactual images remain above baseline. Second, grayscale preserves object-correlated structure, such as texture, material, and luminance layout. We do not treat this structure as a confound: our claim is not that canonical color can be decoded independently of object identity, but that it can be linearly decoded through object-level representations, as examined in Section~\ref{sec:result_object_relation}. %

\subsection{Canonical Color Decodability Across Encoders}
\label{sec:result_gray_histeq}

\begin{figure}[t]
    \centering
    \includegraphics[width=1.0\linewidth]{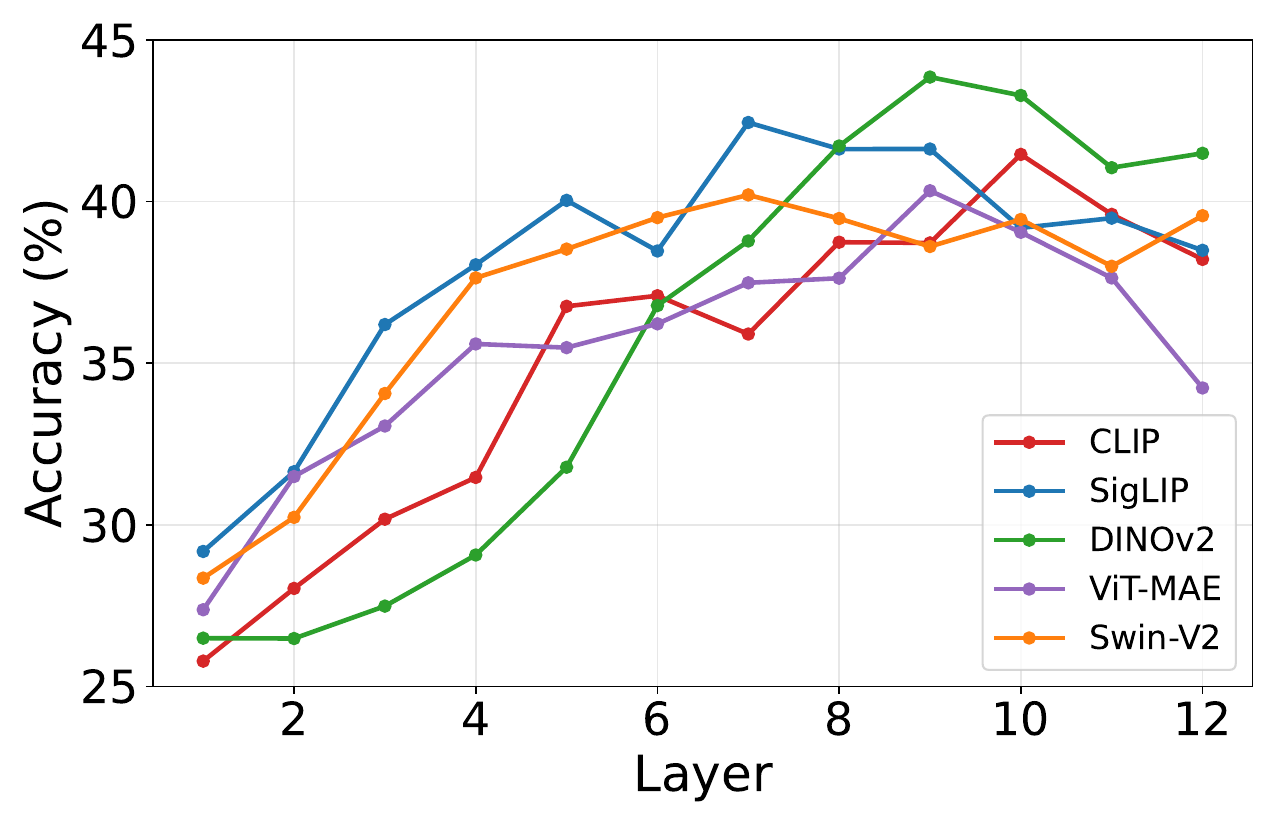}
    \caption{
    Layer-wise color probing accuracy under grayscale inputs using patch-averaged representations.
    The figure compares five vision encoders: CLIP, SigLIP, DINOv2, ViT-MAE, and Swin-V2.
    }
    \label{fig:color_probe_patchavg_gray_histeq}
    \vspace{-1.5em}
\end{figure}

We next ask whether canonical color remains linearly decodable across different vision encoders after visible color cues have been removed. Figure~\ref{fig:color_probe_patchavg_gray_histeq} shows that all five encoders remain above chance and the 13.98\% majority-class baseline under grayscale inputs. Most models improve from early to middle layers, suggesting that canonical-color-predictive information becomes more accessible in later representations. Detailed best-layer results, confidence intervals, and significance tests are reported in Table~\ref{tab:encoder_statistics} in Appendix~\ref{app:statistical_results}.

Across encoders, differences in peak accuracy are modest. DINOv2 reaches the highest overall peak, while SigLIP and Swin-V2 also perform strongly in the middle layers. CLIP improves across layers but does not clearly outperform the self-supervised encoders, suggesting that the signal does not depend exclusively on image--text supervision. Several models, especially ViT-MAE and to a lesser extent SigLIP and DINOv2, show some degradation in their final layers, which may reflect increasing specialization for their pretraining objectives.

Because this cross-encoder comparison holds model scale approximately fixed, we additionally conduct a controlled within-family comparison of DINOv2-S/B/L/g using the same probing protocol.

\begin{table}[t]
\centering
\small
\resizebox{\linewidth}{!}{
\begin{tabular}{@{}lcccc@{}}
\toprule
\textbf{Model} & \textbf{Params} & \textbf{RGB} & \textbf{Gray} & \textbf{Gray + HE} \\
\midrule
DINOv2-S & 22M  & 71.1 & 46.2 & 41.8 \\
DINOv2-B & 86M  & 77.0 & 48.6 & 43.8 \\
DINOv2-L & 304M & 79.4 & 51.0 & 46.0 \\
DINOv2-g & 1.1B & 80.6 & 51.8 & 48.0 \\
\bottomrule
\end{tabular}
}
\caption{
Best-layer canonical-color accuracy (\%) across different DINOv2 model scales. All results are obtained using the same object-class-level five-fold protocol and mean-pooled patch features as in the main experiments. HE denotes histogram equalization.
}
\label{tab:dinov2_scale_color}
\vspace{-1.0em}
\end{table}

Canonical-color accuracy increases monotonically with scale in all three input settings (Spearman's $\rho=1.00$): from 46.2\% to 51.8\% for Gray and from 41.8\% to 48.0\% for Gray+HE. The RGB--Gray gap remains broadly stable across sizes, while entity CLS accuracy follows the same monotonic trend. Thus, canonical-color decodability under grayscale inputs is not specific to a single Base-scale operating point and becomes stronger with model scale.

Together, the cross-encoder and scale comparisons establish that canonical color remains linearly accessible after visible color cues are removed. They do not, however, determine whether this accessibility is mediated by object-level recognition; the next section examines this relationship directly.

\subsection{Relation to Object-Class Recognition}
\label{sec:result_object_relation}

\begin{figure}[t]
    \centering
    \includegraphics[width=1.00\linewidth]{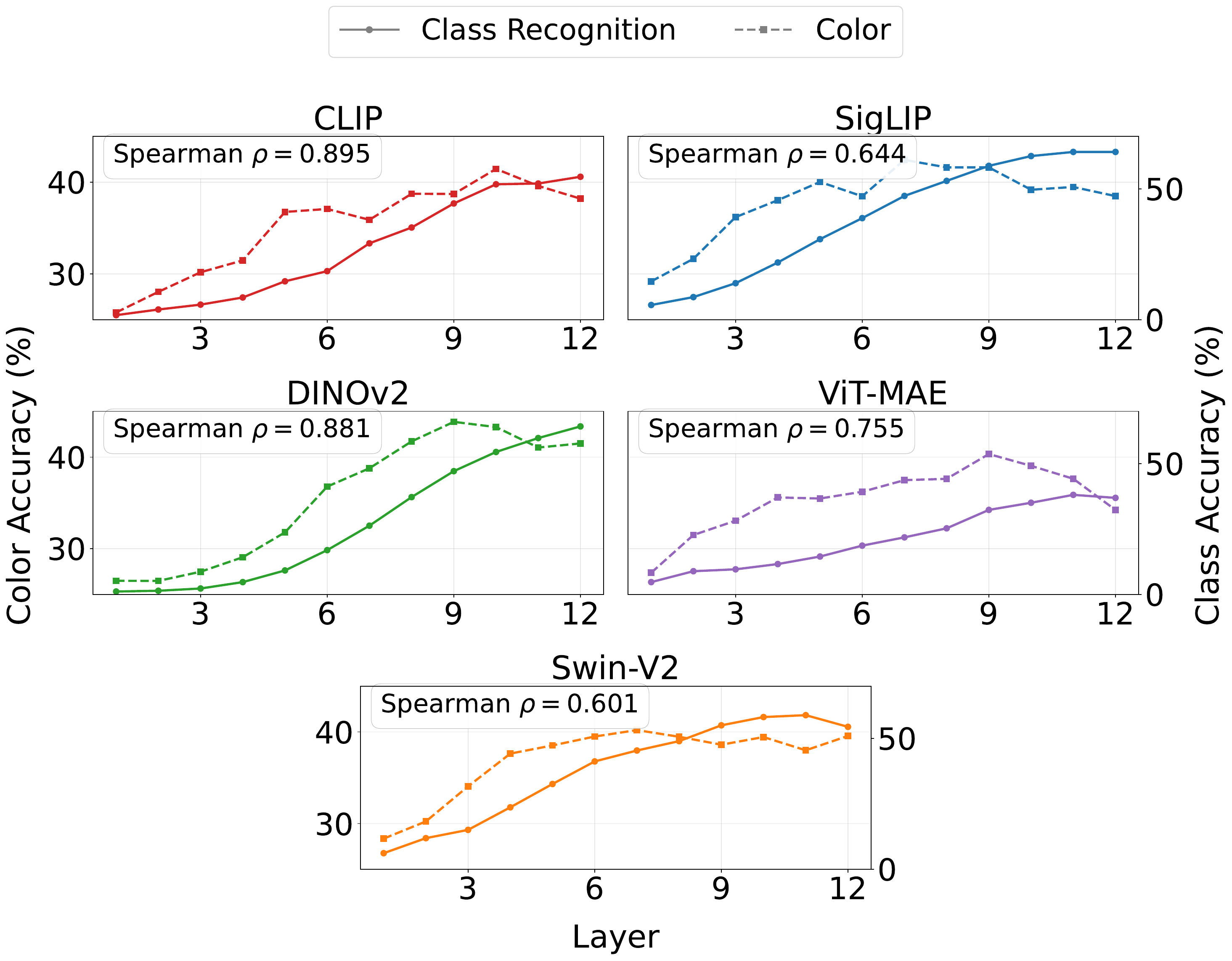}
    \caption{
    Layer-wise object-class recognition and color probing under grayscale inputs.
    Solid and dashed lines show object-class and color probing accuracy, respectively.
    Each subplot reports the Spearman correlation between the two layer-wise curves.
    }
    \label{fig:patchavg_object_recognition_vs_color_probe}
    \vspace{-1.0em}
\end{figure}

We analyze how color probe and object recognition probe results relate. Both probes are trained on the same patch-averaged representations at each layer, under the grayscale setting. Figure~\ref{fig:patchavg_object_recognition_vs_color_probe} shows that object-class recognition improves with depth for all models, although ViT-MAE remains lower in absolute accuracy. Color probing also tends to improve from early to middle layers, but its layer-wise trajectory does not always match object-class recognition. For example, some models show high color accuracy before object recognition reaches its peak, while others show a late-layer drop in color probing despite continued gains in object recognition. This suggests that color-predictive information is associated with object-level representations, but is not fully determined by object-class recognition accuracy alone.

We measure Spearman correlation as a measure of layer-wise association between the two accuracy curves (shown in Figure~\ref{fig:patchavg_object_recognition_vs_color_probe}), to establish whether layers with higher object-class recognition accuracy also tend to have higher color probing accuracy. All models show a strong positive correlation, with CLIP and DINOv2 having the strongest layer-wise alignment. This suggests that color-predictive information and object-class information become more accessible in similar parts of the encoder, although the two signals are not perfectly aligned.

\begin{figure}[t]
    \centering
    \includegraphics[
        width=1.00\linewidth,
        trim=40 0 0 0,
        clip
    ]{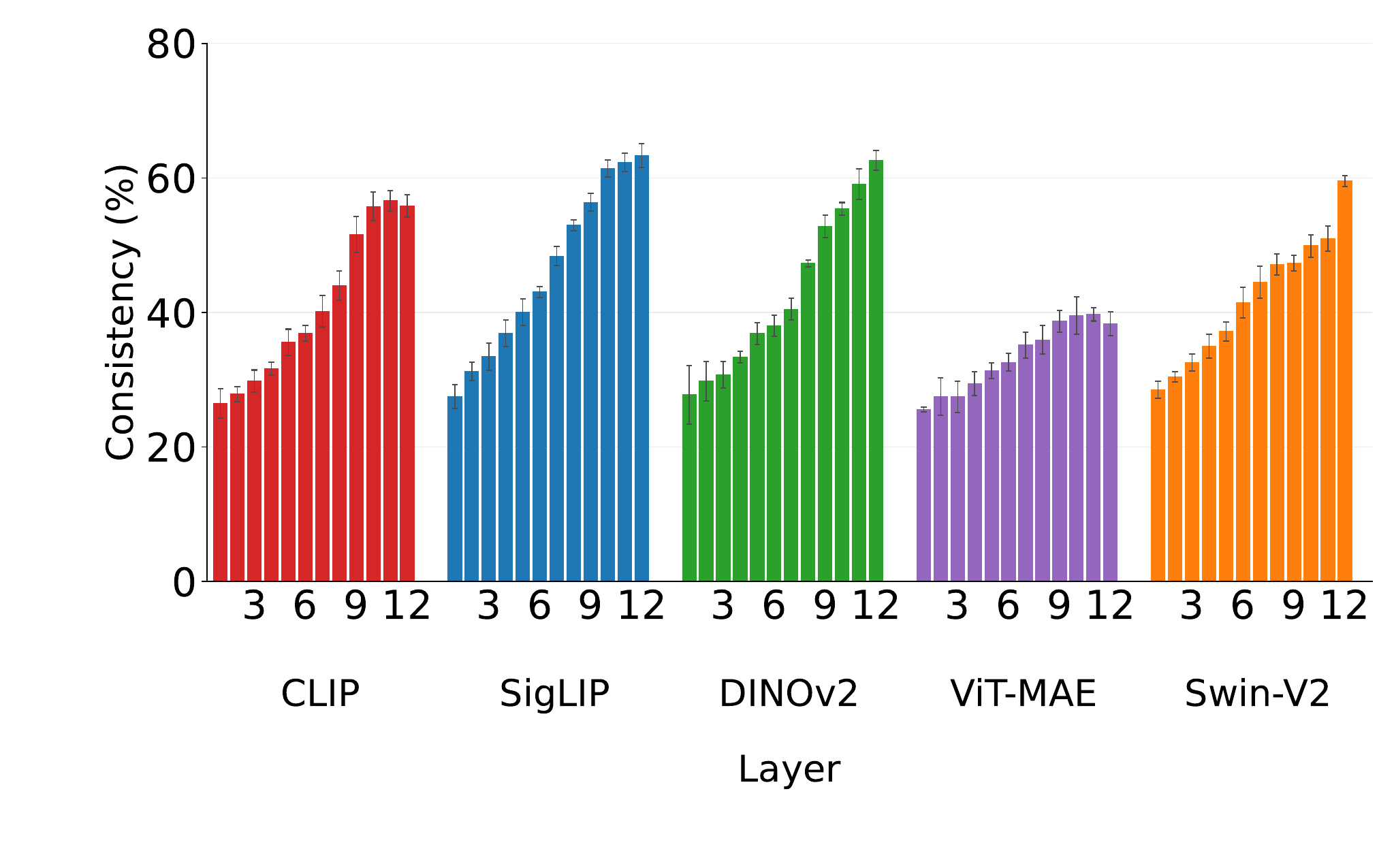}
    \caption{
    Predicted-object color accuracy under grayscale inputs.
    Bars report the percentage of test entities for which the color probe predicts the canonical color of the object predicted by the object probe.
    Error bars show $\pm 1$ standard deviation across five folds.
    }
    \label{fig:predicted_object_color_accuracy}
    \vspace{-1.0em}
\end{figure}

Figure~\ref{fig:predicted_object_color_accuracy} measures color accuracy under the object class assumed by the object probe. 
For each test entity, let $\hat{o}$ be the object predicted by the object probe, $\hat{c}$ be the color predicted by the color probe, and $g(\hat{o})$ be the canonical color of the predicted object. 
We count the prediction as correct when $\hat{c}=g(\hat{o})$. 
This metric does not require $\hat{o}$ to match the ground-truth object; it asks whether the color prediction is correct relative to the model's own object prediction.

The score increases in later layers for most encoders, reaching around 55--65\% for CLIP, SigLIP, DINOv2, and Swin-V2, with ViT-MAE lower. 
This provides stronger evidence that the color probe often tracks object-level canonical color information, rather than behaving independently of object identity. 
At the same time, the scores are not near ceiling, which suggests that the encoders do not expose canonical color knowledge equally well for all object classes. 

\section{VLM Post-training and the Location of Semantic Decodability}
\label{sec:vlm_semantic_decodability}

We next ask whether VLM post-training changes where canonical color and object information are linearly decodable. We do not evaluate final VLM answers, since they can be confounded by decoder-side world knowledge. Instead, we probe frozen representations across matched pre-/post-VLM vision towers and decoder-side visual-token states.

\subsection{Model-dependent changes in the vision tower}
\label{subsec:vlm_vision_towers}

We first test whether post-training changes linear accessibility in the standalone vision tower. We compare three matched pre-/post-VLM pairs: OpenAI CLIP ViT-L/14@336px versus the CLIP vision tower in Molmo~\citep{deitke2025molmo}, OpenAI CLIP ViT-L/14@224 versus the CLIP vision tower in mPLUG-Owl~\citep{ye2023mplug}, and SigLIP-So400m@224 versus the SigLIP vision tower in PaliGemma~\citep{beyer2024paligemma}. Within each pair, the pre- and post-VLM encoders are evaluated at the input resolution used by the VLM; across pairs, we compare the direction of change rather than absolute accuracy. Model repositories are provided in Appendix~\ref{sec:appendix}.

All models are evaluated under the same grayscale setting. For each model and layer, we extract patch-averaged features and train linear probes on frozen representations. Canonical-color probing uses object-class-level cross-validation, while object recognition uses the same image-level entity-recognition protocol as before.

\begin{figure*}[t]
    \centering
    \includegraphics[width=\textwidth]{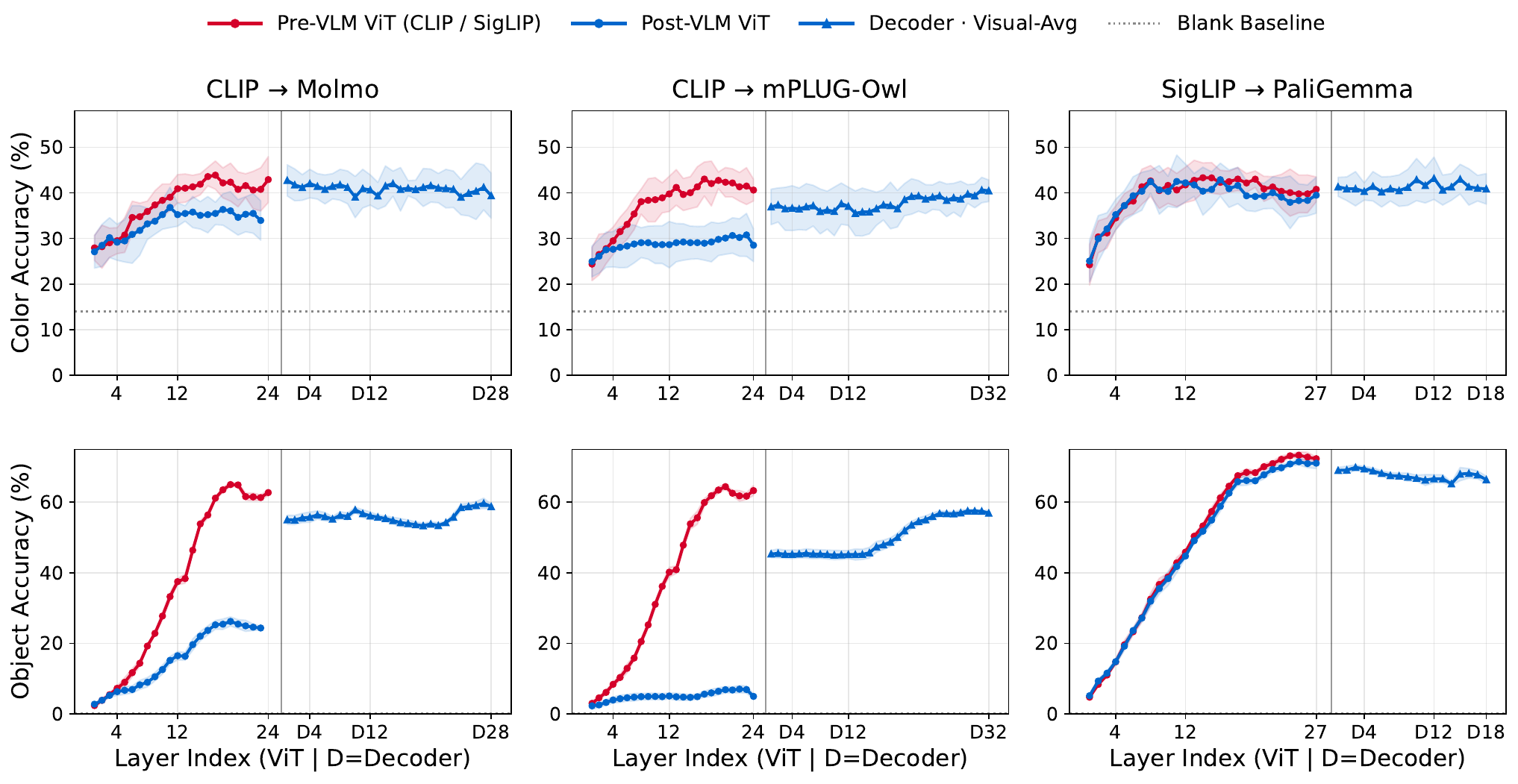}
    \caption{
    Probing semantic decodability across vision towers and decoder visual-token states under grayscale inputs.
    Columns show matched pre-/post-VLM pairs: CLIP--Molmo, CLIP--mPLUG-Owl, and SigLIP--PaliGemma.
    Top row: canonical-color probing.
    Bottom row: object recognition.
    Red circles show the pre-VLM vision encoder.
    Blue circles show the post-VLM vision tower.
    Blue squares show the decoder \textit{visual-avg} readout.
    Dotted gray lines show blank-image baselines; shaded bands show \(\pm 1\) standard deviation across folds.
    }
    \label{fig:vlm_stack_probe}
    \vspace{-1.0em}
\end{figure*}

The red- and blue-circle curves in Figure~\ref{fig:vlm_stack_probe} show that post-training has model-dependent effects on vision-tower decodability. In the two CLIP-based pairs, Molmo and mPLUG-Owl, the post-VLM vision towers yield lower linear-probe accuracy than the corresponding original CLIP encoders. The reduction is moderate for canonical color but substantially larger for object recognition. Thus, in these two models, post-training makes object-level information less linearly accessible from the standalone vision tower.

The SigLIP--PaliGemma pair behaves differently. Under the same grayscale probing setup, the post-VLM PaliGemma vision tower closely follows the original SigLIP encoder for both canonical color and object recognition. Reduced standalone vision-tower decodability is therefore not a universal consequence of VLM post-training: the two CLIP-based VLMs show clear reductions, whereas PaliGemma largely preserves the original SigLIP pattern. Detailed numerical comparisons and paired significance tests are reported in Table~\ref{tab:vlm_statistics} in Appendix~\ref{app:statistical_results}.

\subsection{Visual-interface and decoder-side probing}
\label{subsec:decoder_side_probing}

To test the reformatting hypothesis, we ask whether information weakly decodable from the post-VLM vision tower becomes linearly accessible again downstream. Keeping each VLM frozen, we run the same grayscale images through the model, cache hidden states after the visual interface and at each decoder layer, and apply the same linear probes as before. At each decoder layer, we average the hidden states at the visual-token positions passed to the language model, yielding the \textit{visual-token average} (\textit{visual-avg}), the closest decoder-side analogue of the patch-averaged vision features. Decoder layers \(d_1,\ldots,d_N\) are plotted after the vision layers only to indicate where information becomes linearly accessible in the full stack. A blank-image baseline tests whether the recovered signal depends on visual input.

The blue-square curves in Figure~\ref{fig:vlm_stack_probe} show that object and canonical-color information can become linearly accessible again in decoder-side representations. In Molmo, canonical-color decodability returns close to the original CLIP level from the first decoder layer, while object recognition increases toward the original CLIP peak in later layers. mPLUG-Owl shows the same pattern: despite weak standalone-tower decodability, its decoder-side representations recover much of the reduction for both tasks. PaliGemma behaves somewhat differently: its post-VLM vision tower already closely tracks the original SigLIP encoder, so its decoder-side readout remains broadly comparable rather than showing a substantial recovery.

To locate where the recovery enters the stack, we additionally probe the mPLUG-Owl visual interface for object recognition.

\begin{figure}[t]
    \centering
    \includegraphics[
        width=1.00\linewidth,
        trim=40pt 90pt 0pt 10pt,
        clip
    ]{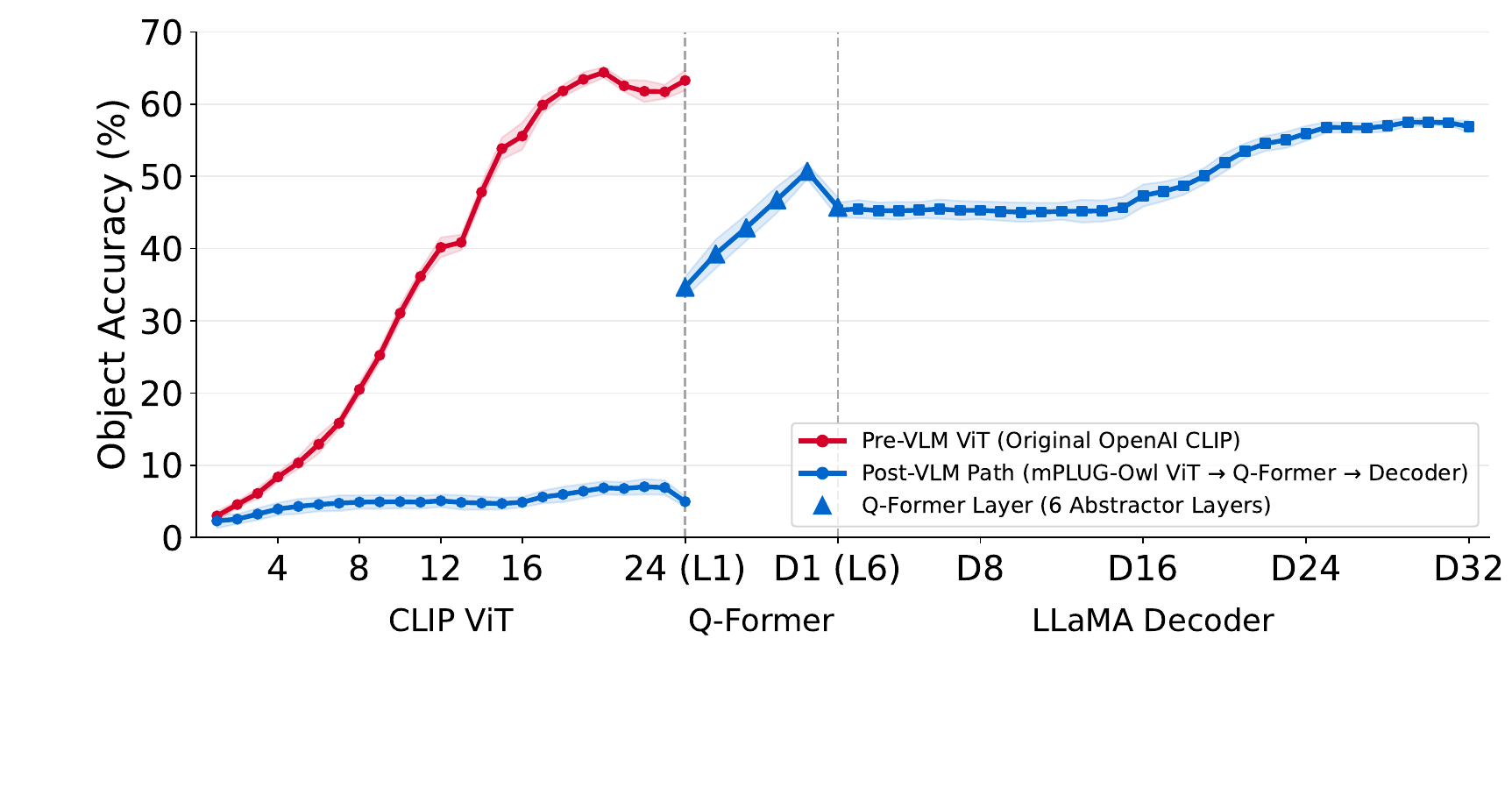}
    \caption{
    Object-recognition probing along the mPLUG-Owl visual path under grayscale inputs.
    Red shows the original CLIP encoder; blue shows the post-VLM path through the visual abstractor and LLaMA decoder.
    The main increase occurs at the visual interface.
    }
    \label{fig:mplug_visual_interface_probe}
    \vspace{-1.0em}
\end{figure}

Figure~\ref{fig:mplug_visual_interface_probe} shows that the post-VLM ViT readout remains low, whereas accuracy is already substantially higher at the Q-Former-style visual abstractor output, before the visual tokens enter the LLM decoder. The first decoder layer changes this value only slightly, while later layers further refine it, suggesting that the main increase occurs at the visual interface rather than in the language decoder alone.

This pattern reflects a change in linear accessibility, not proof that the post-VLM ViT contains no object information; a nonlinear probe might recover additional information. We focus on information directly accessible to a simple linear readout to remain consistent with the earlier experiments. The blank-image baseline remains near the majority-class baseline, suggesting that the recovered signal depends on visual input.

Overall, these results are consistent with reformatting rather than removal. In Molmo and mPLUG-Owl, object and canonical-color information becomes less linearly accessible in the standalone tower but reappears after the learned visual interface and in decoder-side representations. PaliGemma, whose vision tower changes less substantially, largely preserves the original decodability pattern.

\subsection{Prompt-level behavior under color-controlled inputs}
\label{subsec:vlm_prompt_check}

As a final behavioral check, we query the VLMs with color- and object-related prompts. This experiment is not intended to localize color information; instead, it illustrates how final answers can mix visible evidence, object recognition, canonical-color priors, and output formatting.

We evaluate the same 708 object classes after converting all images to grayscale. We use three prompts: \textit{Visual Color}, ``What color is the object in this image? Answer with just one word,'' for which the expected answer is gray; \textit{Canonical Color}, ``What is the typical real-world color of this object? Answer with just one word,'' which targets typical rather than visible color; and \textit{Object}, ``What object is shown in this image? Answer with one or two words,'' which tests object identification.

For the two color prompts, we obtain class-level predictions by majority vote over images of the same class. Because exact matching to Wikidata names is too strict for fine-grained, scientific, and proper names, we evaluate object recognition using manually verified acceptable labels derived from the entity name and Wikidata description. VLM outputs are not used to construct these label sets.

\begin{table}[t]
\centering
\small
\resizebox{\linewidth}{!}{
\begin{tabular}{@{}lccc@{}}
\toprule
\textbf{Metric} & \textbf{Molmo} & \textbf{mPLUG} & \textbf{PaliGemma} \\
\midrule
Visual color & 23.4 & 14.9 & 17.3 \\
Canonical color & 44.5 & 25.6 & 4.1 \\
Object category & 53.0 & 57.1 & 61.3 \\
\midrule
Dec. color probe & 42.7 & 40.6 & 43.2 \\
Dec. object probe & 59.7 & 57.5 & 69.9 \\
\bottomrule
\end{tabular}
}
\caption{
Prompt-level results on grayscale images.
The first three rows report VLM answer accuracy; the last two report peak decoder \textit{visual-avg} probe accuracy.
}
\label{tab:prompt_level_check}
\vspace{-1.0em}
\end{table}

Table~\ref{tab:prompt_level_check} shows that final answers are not a clean measure of visual color perception. All models perform poorly on \textit{Visual Color} despite gray being the expected answer, suggesting effects from object-level color priors or response preferences. \textit{Canonical Color} improves performance substantially for Molmo and also for mPLUG-Owl, while PaliGemma performs poorly. Category-level object recognition is higher but remains imperfect under grayscale inputs. Thus, final answers conflate visual evidence, object recognition, priors, and output normalization, whereas internal visual-token probes provide a cleaner view of where object and canonical-color information are linearly decodable.

\section{Conclusion}

Using canonical color as a controlled probe, we find that it remains linearly decodable from grayscale images even though RGB probing is strongly influenced by visible color. Counterfactual and object probes show that this signal aligns more closely with canonical than surface color and is related to, but not fully determined by, object identity. VLM post-training can reduce tower-level decodability while the information reappears after the visual interface or within decoder-side states, a pattern consistent with representational reformatting rather than removal and illustrating how canonical color information traces object-level semantics across the vision–language stack.

\section*{Limitations}

Our study focuses on canonical color as a controlled case study. Color is useful because visible chromatic cues can be removed with grayscale conversion and histogram equalization, but this also means that our empirical conclusions should not be directly generalized to all visual attributes. Other attributes, such as material, texture, function, or affordance, may require different controls and evaluation protocols.

Our analysis is also based on linear probing. This allows us to compare representations across models and layers in a simple and consistent way, but it only measures information that is linearly accessible from the chosen readout. A nonlinear probe or a downstream task-specific head might recover additional information. Therefore, our results should be interpreted as evidence about linear decodability, rather than as a complete measure of all information contained in the representations.

\section*{Acknowledgments}
Stella Frank was funded by NNF project 0094281.

\bibliography{custom}

\clearpage
\newpage

\appendix

\section{Additional Dataset and Model Details}
\label{sec:appendix}

This appendix provides additional details on dataset construction, label sources, evaluation splits, and the model repositories used in our experiments.

\subsection{Dataset Construction and Label Source}
\label{sec:appendix_dataset}

The dataset contains 708 object concepts. Canonical-color labels are obtained from the Wikidata color property (\texttt{P462}). We retain only entities associated with a single color value and map the source values to ten basic categories: black, blue, brown, gray, green, pink, purple, red, white, and yellow. This reduced label space makes the probing task tractable and allows direct comparison across object classes. The canonical-color labels are therefore derived from structured Wikidata entries rather than assigned through free-form human annotation.

For each object concept, we collect images linked from its Wikidata entry and supplement them with Google Search results, with a target of five images per concept. Manual inspection is used only for image quality control and does not assign or revise the canonical-color labels. We remove images that do not clearly depict the target object or are inconsistent with its canonical color. Because filtering is performed after image collection, some concepts contain fewer than five images.

\subsection{Class Distribution and Sampling}

We retain the resulting imbalance across canonical-color labels rather than forcing all categories to have the same size. During collection, Pok\'emon formed a large subset of the candidate concepts. We therefore cap this subset at 50 concepts while prioritizing the less frequent gray, brown, and red categories. The final dataset contains 69 black, 69 blue, 53 brown, 48 gray, 73 green, 81 pink, 65 purple, 61 red, 99 white, and 90 yellow concepts. White is the most frequent label and gray is the least frequent.

\begin{figure}[t]
    \centering
    \includegraphics[width=\linewidth]{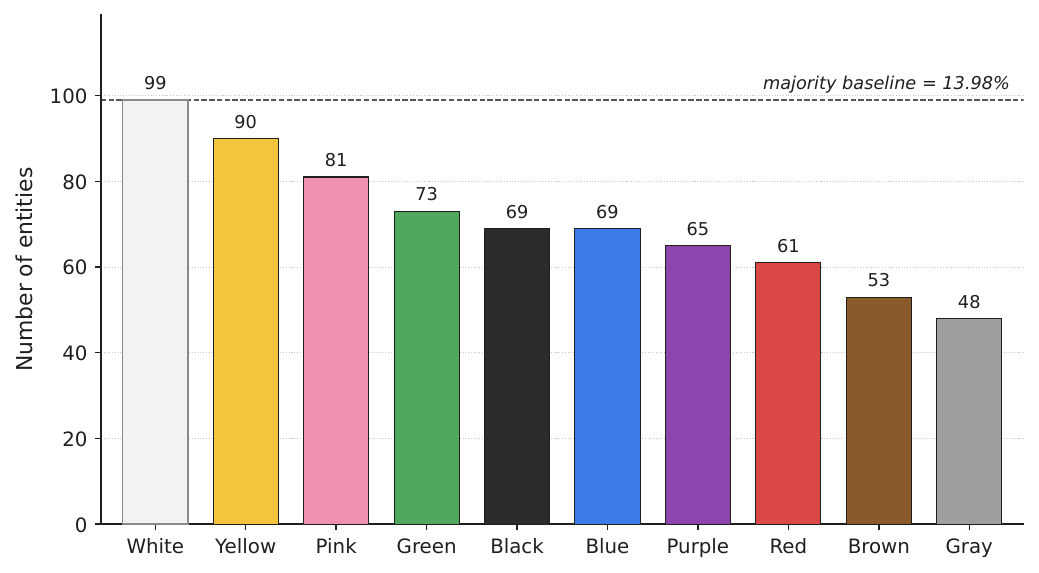}
    \caption{
    Distribution of the 708 object concepts across the ten canonical-color labels.
    }
    \label{fig:color_distribution}
\end{figure}

\subsection{Evaluation Splits and Variable Image Counts}

For color probing, we use object-class-level five-fold cross-validation. All images of the same object concept are assigned to the same fold. In each run, the probe is trained on four folds and evaluated on object concepts from the held-out fold. The test classes are therefore unseen during training, so this setting evaluates canonical-color generalization to new object concepts rather than to new images of known concepts.

At test time, we average the predicted color-probability vectors over all available images of each object concept and take the class with the highest average probability. This produces one color prediction per object concept regardless of its number of images. We report results against the approximately 14\% majority-class baseline. The cross-validation procedure defines the held-out object concepts but does not rebalance the color-label distribution.

For object recognition, we instead use a within-class image hold-out because holding out an entire object class would remove its target label from training. For each concept with at least two images, we hold out one image for testing and use the remaining images for training. Concepts represented by a single image are used for training only.

\subsection{Model Repositories}

\begin{table*}[t]
\centering
\small
\resizebox{\textwidth}{!}{
\begin{tabular}{lll}
\toprule
\textbf{Role} & \textbf{Model} & \textbf{Hugging Face repository} \\
\midrule
Vision encoder & CLIP ViT-B/32 & \url{https://huggingface.co/openai/clip-vit-base-patch32} \\
Vision encoder & SigLIP Base & \url{https://huggingface.co/google/siglip-base-patch16-224} \\
Vision encoder & DINOv2 Base & \url{https://huggingface.co/facebook/dinov2-base} \\
Vision encoder & ViT-MAE Base & \url{https://huggingface.co/facebook/vit-mae-base} \\
Vision encoder & Swin-V2 Base & \url{https://huggingface.co/microsoft/swinv2-base-patch4-window12-192-22k} \\
\midrule
Pre-VLM vision tower & CLIP ViT-L/14@336px & \url{https://huggingface.co/openai/clip-vit-large-patch14-336} \\
Pre-VLM vision tower & CLIP ViT-L/14@224px & \url{https://huggingface.co/openai/clip-vit-large-patch14} \\
Pre-VLM vision tower & SigLIP-So400m@224px & \url{https://huggingface.co/google/siglip-so400m-patch14-224} \\
\midrule
VLM & Molmo-7B-D & \url{https://huggingface.co/allenai/Molmo-7B-D-0924} \\
VLM & mPLUG-Owl & \url{https://huggingface.co/MAGAer13/mplug-owl-llama-7b} \\
VLM & PaliGemma 3B Mix 224 & \url{https://huggingface.co/google/paligemma-3b-mix-224} \\
\bottomrule
\end{tabular}
}
\caption{
Hugging Face model repositories used in our experiments.
}
\label{tab:hf_model_urls}
\vspace{-1.0em}
\end{table*}

\section{Numerical Results and Statistical Tests}
\label{app:statistical_results}

This section provides numerical summaries of the layer-wise comparisons in the main text. All results use histogram-equalized grayscale inputs and five folds. For the encoder comparison, we report best-layer accuracy with 95\% confidence intervals, one-sample $t$-tests against the 13.98\% majority baseline, and paired $t$-tests for the RGB-to-grayscale difference. For the matched VLM pairs, we report best-layer pre-/post-training differences and paired $t$-tests over the five folds.

\paragraph{Vision encoders.}
Table~\ref{tab:encoder_statistics} reports the best-layer canonical-color results for the five vision encoders. All grayscale accuracies are significantly above the majority baseline, while all RGB-to-grayscale reductions are also significant.

\begin{table}[t]
\centering
\small
\setlength{\tabcolsep}{3pt}
\resizebox{\columnwidth}{!}{
\begin{tabular}{lcccc}
\toprule
Encoder &
Gray Acc. &
95\% CI &
vs.\ 13.98\% &
RGB$\rightarrow$Gray Drop \\
\midrule
CLIP    & 41.5\% & [38.4, 44.6] & $p<.001$ & 34.9 pp ($p<.001$) \\
DINOv2  & 43.8\% & [37.0, 50.7] & $p<.001$ & 32.8 pp ($p<.001$) \\
SigLIP  & 42.4\% & [37.7, 47.2] & $p<.001$ & 39.5 pp ($p<.001$) \\
ViT-MAE & 40.3\% & [35.3, 45.4] & $p<.001$ & 35.3 pp ($p<.001$) \\
Swin-V2 & 40.2\% & [35.8, 44.6] & $p<.001$ & 37.1 pp ($p<.001$) \\
\bottomrule
\end{tabular}
}
\caption{Best-layer canonical-color probing under histogram-equalized grayscale inputs. Confidence intervals are computed over five folds. Significance relative to the 13.98\% majority baseline is assessed using one-sample $t$-tests; RGB-to-grayscale differences use paired $t$-tests. Here, pp denotes percentage points.}
\label{tab:encoder_statistics}
\end{table}

\paragraph{Pre-/post-VLM vision towers.}
Table~\ref{tab:vlm_statistics} compares the standalone vision towers before and after VLM post-training. The two CLIP-based pairs show substantial reductions, particularly for object recognition, whereas the SigLIP--PaliGemma pair changes little in absolute magnitude.

\begin{table}[t]
\centering
\small
\setlength{\tabcolsep}{3pt}
\resizebox{\columnwidth}{!}{
\begin{tabular}{llrrrr}
\toprule
VLM Pair & Task & Pre-VLM & Post-VLM & Change & Paired Test \\
\midrule
CLIP$\rightarrow$Molmo
    & Color  & 43.9\% & 36.8\% & $-7.1$ pp  & $p<.05$  \\
CLIP$\rightarrow$Molmo
    & Object & 65.0\% & 26.2\% & $-38.8$ pp & $p<.001$ \\
CLIP$\rightarrow$mPLUG-Owl
    & Color  & 43.0\% & 30.8\% & $-12.3$ pp & $p<.05$  \\
CLIP$\rightarrow$mPLUG-Owl
    & Object & 64.4\% & 7.0\%  & $-57.4$ pp & $p<.001$ \\
SigLIP$\rightarrow$PaliGemma
    & Color  & 43.3\% & 42.9\% & $-0.4$ pp  & n.s.     \\
SigLIP$\rightarrow$PaliGemma
    & Object & 73.3\% & 71.5\% & $-1.9$ pp  & $p<.001$ \\
\bottomrule
\end{tabular}
}
\caption{Best-layer grayscale probing accuracy for matched standalone vision towers before and after VLM post-training. Significance is assessed using paired $t$-tests over five folds. Statistical significance should be interpreted together with effect magnitude: the PaliGemma object difference is consistent across folds but only 1.9 percentage points.}
\label{tab:vlm_statistics}
\end{table}

\end{document}